\pdfoutput=1
\documentclass[11pt]{article}
\usepackage{coling2020}\colingfinalcopy
\usepackage{times}
\usepackage{url}
\usepackage{latexsym}
\usepackage{hyperref}
    \hypersetup{
        colorlinks=true,    
        linkcolor=blue,     
        citecolor=blue,     
        filecolor=blue,     
        urlcolor=blue
    }
\usepackage{amsmath, amssymb, amsthm, mathtools}
\usepackage{xcolor}
\usepackage{graphicx}
\usepackage{subcaption}
\usepackage{ragged2e, textcomp}
\usepackage{titlesec}
\usepackage{fancyvrb}
\titleformat{\paragraph}{\normalfont\normalsize\bfseries}{\theparagraph}{1em}{}
\titlespacing*{\paragraph}{0pt}{3.25ex plus 1ex minus .2ex}{1.5ex plus .2ex}
\newtheorem*{theorem*}{Theorem}

\newcommand{\numberthis}{\addtocounter{equation}{1}\tag{\theequation}}
\newcommand{\conceptone}[1]{\textcolor{blue}{#1}}
\newcommand{\concepttwo}[1]{\textcolor{red}{#1}}
\title{
    \textbf{
        Revisiting the Prepositional-Phrase Attachment Problem \\ Using Explicit Commonsense Knowledge
    }
}
\author{
    Yida Xin \\
    {\tt yxin@bu.edu} \\ \And 
    Henry Lieberman \\
    {\tt lieber@media.mit.edu} \\ \And 
    Peter Chin \\
    {\tt spchin@cs.bu.edu}
}
\date{}

\begin{document}
\maketitle

\begin{abstract}
    We revisit the challenging problem of resolving prepositional-phrase (PP) attachment ambiguity.\, 
    To date, proposed solutions are either \textit{rule-based}, where explicit grammar rules direct how to resolve ambiguities; or \textit{statistical}, where the decision is learned from a corpus of labeled examples \cite{Hamdan2018AnAO}.\, 
    We argue that explicit commonsense knowledge bases can provide an essential ingredient for making good attachment decisions.\, 
    We implemented a module, named \textsc{Patch-Comm}, that can be used by a variety of conventional parsers, to make attachment decisions.\, Where the commonsense KB does not provide direct answers, we fall back on a more general system that infers ``out-of-knowledge-base'' assertions in a manner similar to the way some NLP systems handle out-of-vocabulary words.\,
    Our results suggest that the commonsense knowledge-based approach can provide the best of both worlds, integrating rule-based and statistical techniques.\, As the field is increasingly coming to recognize the importance of explainability in AI, a commonsense approach can enable NLP developers to better understand the behavior of systems, and facilitate natural dialogues with end users.
\end{abstract}

\section{Introduction}
Many applications and social scenarios nowadays are eager to see reliable and trustworthy Artificial Intelligence (AI) systems.\, But, in order to really establish trust, it is not enough to have computers only be able to communicate the end results of their ``thought process.''\, Instead, computers, from the ground up, should be expected to use \textit{explainable} processes to obtain their results, because that makes it easier --- and in some cases, possible at all --- for people to think about what reasoning processes the computers must have employed to get to those results.\, An even better solution is to build computers that can actually tell us, in natural language, what their reasoning and thinking processes are.
\par
Today's neural-network based language models (LM) --- such as Transformers and its descendants \cite{vaswani2017attention,devlin2018bert,liu2019roberta} --- learn implicit knowledge from enormous corpora, but their representations, such as neural network weights, may only make sense to the systems themselves.\, Work has been done on constructing \emph{post-facto} explanations for opaque inference processes \cite{Petroni2019LanguageMA,da-kasai-2019-cracking,shwartz2020unsupervised}, but it is questionable how general the models and explanations really are.\, They must grapple with a tradeoff between faithfulness to the original process, and understandability by people.\, On the other hand, if we build a language-understanding system based on explicit commonsense knowledge and inference, from the ground up, reporting the commonsense involved is a good way to provide explanation to users.\, But it is hard to acquire, explicitly, sufficient knowledge to achieve wide coverage.\, So knowledge acquisition needs to be automated, which requires the use of Machine Learning mechanisms.
\par
LMs have been proven exceptionally powerful at what they do.\, Recently, more empirical findings have validated our intuition about these systems: They are great knowledge learners for the domains that they are evaluated in \cite{Sakaguchi2019WINOGRANDEAA,shwartz2020unsupervised}.\, It seems unwarranted not to consider using their great power, just because they are somewhat opaque.
\par
That leads us to the statement of our research agenda: We make language understanding explainable by using explicit commonsense knowledge, and we make the knowledge acquirable by using Machine Learning mechanisms, even if they may be somewhat opaque.
\par
As a first step, we focus on a small but notoriously challenging problem in the realm of language understanding: the prepositional-phrase (PP) attachment problem \cite{Hamdan2018AnAO}.\, We also focus on a particular knowledge base and a particular neural-network based architecture for extending the knowledge contained in that knowledge base.

\section{Background and Related Work}
Hamdan and Khan \cite{Hamdan2018AnAO} wrote a comprehensive survey that highlights many historical attempts at the PP attachment problem.\, Very roughly speaking, there are two approaches: (1) PP attachment can either be treated as a classification/ranking problem where one classifies/ranks the place in the sentence that makes the most sense to associate with the said PP; (2) PP attachment can also be dealt with by writing specific grammars and rules for a parser, where such grammars and rules tell the parser where, by default, the PP is to be attached within the sentence.
\par
In this paper, we work with a rule-based parser that has its own default way of always attaching PPs to the highest \textit{head} in the parse tree.\, The parser is part of a question-answering system, and that heuristic was deemed to be often appropriate in that application.\, But recognizing that the default way is not always desirable, we implement a system that makes informed decisions for the parser, so that the parser can consult with our system.\, And by \textit{informed}, we mean that we have our module querying for knowledge from a knowledge base.


\subsection{The START Parser}
The parser we work with is the START parser \cite{Katz1997AnnotatingTW}.\, The first reason that we use START is that it is a syntactic parser, whose nested-triples representation allows for additions of semantic knowledge.\, As an example of how the START parser works, suppose we request START to parse the sentence
\begin{align*}
    \text{``John is playing the guitar with one finger,''} \numberthis{\label{}}
\end{align*}
START will output
\begin{verbatim}
                    [John         play        guitar]
                    [play         with        finger]
                    [finger   has_quantity      1   ]
\end{verbatim}
where \texttt{[John play guitar]} and \texttt{[play with finger]} are the results of syntactic knowledge, and \texttt{[finger has\_quantity 1]} is semantic knowledge.
\par
The second reason we use the START parser is that it is the parser used by the Genesis story-understanding system \cite{genesis-manifesto-2018}.\, Our long-term research agenda, which we will articulate in section 4, is to scale up the language-processing capability of existing  story-understanding systems like Genesis.\, 


\subsection{The ConceptNet Knowledge Base}
The knowledge base we use is the ConceptNet commonsense knowledge base \cite{speer2017conceptnet}.\, The first reason that we use ConceptNet is that it is very large: It contains crowdsourced ``pure'' commonsense knowledge input directly by people, implicit knowledge from behavior in games, and other sources.  It is then combined with more ``factual'' knowledge from WordNet, Wiktionary, and others. See Speer et al. \cite{speer2017conceptnet}.
\par
The second reason is that there have been successful attempts at using ConceptNet's knowledge to reason about out-of-knowledge concepts, using Machine Learning methods \cite{ColonHernandez2019DoesAD}.\, We would like to take advantage of that research result to give ConceptNet a form of knowledge-acquisition capability, thus advancing our research agenda.
\par
Each piece of knowledge in ConceptNet is represented as an assertion that looke like:
\begin{verbatim}
                  [<relation>, <concept1>, <concept2>]
\end{verbatim}
A concrete example could be:
\begin{verbatim}
                [/r/HasA/, /c/en/guitar/, /c/en/strings/]
\end{verbatim}
where \texttt{HasA} is the relation (\texttt{/r/}) that connects the concept (\texttt{/c/}) \texttt{guitar} to the concept \texttt{strings}, and \texttt{en} means English.\, Note that this assertion can only be interpreted as \texttt{guitar HasA strings} and not as \texttt{strings HasA guitar}.
\par
Since the work on AnalogySpace \cite{Speer2008AnalogySpaceRT}, every assertion in ConceptNet now comes with pre-learned numerical weights that indicate its confidence level.\, In the example here, the assertion \texttt{guitar HasA strings} has a a weight of $2.83$, indicating that ConceptNet is fairly confident that a guitar has strings, provided that $1.0$ is the default value that indicates the mere existence of an assertion in ConceptNet.

\subsection{Problems with Commonsense Knowledge Bases}
Working closely with ConceptNet has helped us think about commonsense knowledge bases --- as well as the notion of \textit{commonsense} itself --- in much depth.


\subsubsection{The Inference Problem}
One problem with ConceptNet is what we call the \textit{inference problem}.\, Here is an illuminating example: The assertion
\begin{verbatim}
          [/r/IsA/, /c/en/guitar/, /c/en/string_instrument/]
\end{verbatim}
exists in ConceptNet, with a weight of $6.32$, which is much higher than the previous $2.83$.\, But, querying ConceptNet with just \texttt{guitar} (as well as its plural form \texttt{guitars}) and \texttt{strings} (as well as its singular form \texttt{string}), we cannot get to \texttt{guitar IsA string\_instrument}, because no assertion that connects \texttt{string} with \texttt{string\_instrument} is explicitly given, and ConceptNet alone does not know how to infer an assertion to connect the two concepts.
\par
The inference problem points to an even more fundamental problem with all commonsense knowledge bases: They merely store knowledge are just data, but they do not know the meanings of those data.\, So, ironically, commonsense knowledge bases lack commonsense.


\subsubsection{The Insufficiency Problem}
The other problem is the \textit{insufficiency problem}, which says that ConceptNet simply does not have all the knowledge in and about the world.\, In fact, all knowledge bases and all corpora suffer from the insufficiency problem, to various degrees.\, To address insufficiency, we utilize the RetroGAN-DRD system, introduced by Colon-Hernandez et al. \cite{ColonHernandez2019DoesAD}, that infers ``out-of-knowledge-base'' assertions in a manner similar to the way some NLP systems handle out-of-vocabulary words.\, We will see that, by using RetroGAN-DRD instead of just ConceptNet, we are guaranteed to always get an answer, which ConceptNet alone cannot guarantee.\, The answers align with our own commonsense and intuition about language, and they inform START's PP attachment decisions in just that manner.
\par
In addition to RetroGAN-DRD, there are such representative works as ATOMIC/COMET \cite{Sap2019ATOMICAA,Bosselut2019COMETCT} in commonsense knowledge base completion and extension.\, There, the Transformers-based neural networks mechanisms learn both to fill in the gaps of knowledge bases (completion) and to generate and add new knowledge into the knowledge bases (expansion).
\par
Both RetroGAN-DRD \cite{ColonHernandez2019DoesAD} and COMET \cite{Bosselut2019COMETCT} achieve a level of \textit{neural-symbolic integration}.\, They ensure that a user interacts with their symbolic, understandable information, and that a developer debugs the underlying Machine Learning mechanisms more easily by using the symbolic information as guidance.


\section{Implementation Details}

We call our system \textsc{Patch-Comm}, and we will use this name for the remainder of this paper.


\subsection{A Simple Case, Using Only ConceptNet}
Suppose we request START to parse the sentence:
\begin{align*}
    \text{``John is playing the guitar with one string.''} \numberthis{\label{}} 
\end{align*}
Then, by default, START alone would output:
\begin{figure}[!ht]
    \centering
    \begin{verbatim}
                [John          play          guitar]
                [play          with          string]
                [strings   has_quantity        1   ]\end{verbatim}
    \caption{START's output for sentence (2), without using \textsc{Patch-Comm}.}
\end{figure}
\par\noindent
Evidently for us, \texttt{guitar}, instead of \texttt{play}, should be the more sensible attachment for \texttt{string}.
So clearly, START needs help, and we will explicitly use ConceptNet's commonsense knowledge to help START to make the more sensible parsing decision.
\par
To do so, we probe START's partial parsing information.\, Specifically, START begins and continues parsing just like before, until it reaches a point of ambiguity.\, But this time, START consults with \textsc{Patch-Comm} for a parsing decision by sending its latest partial parse to \textsc{Patch-Comm}:
\begin{figure}[!ht]
    \centering
    \begin{verbatim}
            (:ambig-PP
                (PP :prep "with" :det "one" :noun "string")
             :possible-attachments
                ((V :verb "play")
                 (NP :det "the" :noun "guitar")))\end{verbatim}
    \caption{START's partial parse for sentence (2).}
\end{figure}
\par
\textsc{Patch-Comm} receives it, and takes out the keyword concepts from both the ambiguous prepositional phrase (\texttt{:ambig-pp}) and from each of the possible attachments.\footnote{A future work is to try to utilize the other information available in this partial parse, such as the preposition itself.\, For example, we can treat the querying of ConceptNet for the most sensible relations between keyword concepts as a first loop; and then, we can build a second loop where \textsc{Patch-Comm} compares the queried relations against the prepositions, and use linguistic information to decide whether, and to what extent, those relations make sense with respect to the prepositions.}\, Namely, \textsc{Patch-Comm} extracts the concept \texttt{string} (and plural \texttt{strings}) as well as the concepts \texttt{play} and \texttt{guitar} (and plural \texttt{guitars}).\footnote{Note that we do not inflect the verb, because in ConceptNet verb concepts are registered in their lemmatized forms.}
\par
We query ConceptNet with a pair of concepts, and ask for the highest-weighted assertion that relates them.\footnote{We did not implement multi-hop searches between the two concepts.\, One reason is that, at the commonsense level, it is hard to know what to make of multiple chained-together assertions with potentially wildly different relations, only to try to justify some sort of connection between those two concepts.\, So we decided to stick with the most straightforward knowledge possible --- just the assertions that already exist.}\, Specifically, for each extracted concept (as well as its lemmatized or inflected version) from the possible attachments, we pair it up with the extracted concept from the ambiguous PP.\, In the example here, \textsc{Patch-Comm} extracts six pairs of concepts, queries ConceptNet with each such pair for a total of six times, and obtains the following information:
\begin{table}[!ht]
    \centering
    \captionsetup{width=0.875\linewidth}
    \begin{tabular}{|c|c|c|}
        \hline
        Concept Pair        & Relation      & Weight \\
        \hline
        (string, play)      & ---           & --- \\
        (strings, play)     & ---           & --- \\
        (\conceptone{string}, \concepttwo{guitar})    & AtLocation    & 2.0 \\
        (string, guitars)   & ---           & --- \\
        (\concepttwo{strings}, \conceptone{guitar})   & HasA          & 2.8284 \\
        (strings, guitars)  & ---           & --- \\
        \hline
    \end{tabular}
    \caption{ConceptNet's symbolic knowledge about the ambiguous PP from sentence (2).\, The concept that appear to the left of a relation is in \conceptone{blue}, and the concept to the right in \concepttwo{red}.}
\end{table}
\newpage
\par
Evidenced by the four empty rows in Table 1, we see that, even after taking into account the singulars and plurals of the nouns, we are still running into the disconnected-knowledge problem (section 2.3.1).\, Still, to agree with lemmatized nouns in the partial parse, we compress Table 1 into Table 2:
\begin{table}[!ht]
    \centering
    \captionsetup{width=0.85\linewidth}
    \begin{tabular}{|c|c|c|}
        \hline
        Concept Pair        & Relation      & Weight \\
        \hline
        (string, play)      & ---           & --- \\
        (\concepttwo{string}, \conceptone{guitar})    & HasA          & 2.8284 \\
        \hline
    \end{tabular}
    \caption{ConceptNet's symbolic knowledge about the ambiguous PP from sentence (2), refined.\, The concept that appear to the left of a relation is in \conceptone{blue}, and the concept to the right in \concepttwo{red}.}
\end{table}
\par\noindent 
Note that we indeed only preserve the highest-weighted relation overall.
\par
Having obtained the information in Table 2, \textsc{Patch-Comm} makes the decision, on behalf of START, that the ambiguous PP (\texttt{ambig-PP} ``with one string'') is to be attached to the noun phrase (\texttt{NP} ``the guitar'') rather than to the verb (\texttt{V} ``play'').\, \textbf{Note, crucially, that this decision contradicts START's default decision.}\, Our decision is then sent back to START, so that START can continue and complete its parse result:
\begin{figure}[!ht]
    \centering
    \begin{verbatim}
                [John          play          guitar]
                [guitar        with          string]
                [string    has_quantity        1   ]\end{verbatim}
    \caption{START's output for sentence (2), using \textsc{Patch-Comm} querying ConceptNet.}
\end{figure}
\par\noindent
Although the only difference between Figure 3 and Figure 1 is changing the word ``play'' to the word ``guitar'' in the second row, it is a big difference.


\subsection{A Less Simple Case, Using Only ConceptNet}
As a more complex example, suppose we have now:
\begin{align*}
    \text{``John is playing the guitar with one string with one finger.''} \numberthis{\label{}}
\end{align*}
What makes sentence (3) more complex is that it has two, instead of one, ambiguous PPs.\, But just like before, we request START to parse sentence (3).\, This time, START pauses as soon as it meets the first ambiguity, and sends \textsc{Patch-Comm} the partial parse:
\begin{figure}[!ht]
    \centering
    \begin{verbatim}
            (:ambig-PP
                (PP :prep "with" :det "one" :noun "finger")
             :possible-attachments
                ((V :verb "play")
                 (NP :det "the" :noun "guitar")
                 (PP :prep "with" :det "one" :noun "string")))\end{verbatim}
    \caption{START's 1st partial parse for sentence (3).}
\end{figure}
\par\noindent
And then, just like before, \textsc{Patch-Comm} extracts concepts, queries ConceptNet with the concepts, and obtains a refined table of information:
\begin{table}[!ht]
    \centering
    \begin{tabular}{|c|c|c|}
        \hline
        Concept Pair        & Relation      & Weight \\
        \hline
        (finger, play)      & RelatedTo     & 1.0 \\
        (finger, guitar)    & ---           & --- \\
        (finger, string)    & ---           & --- \\
        \hline
    \end{tabular}
    \caption{ConceptNet's knowledge about the 1st ambiguous PP from sentence (2), refined.}
\end{table}
\newpage
This time with more blank rows in the table, we are also seeing the insufficient-knowledge problem (section 2.3.2) with ConceptNet, in addition to the disconnected-knowledge problem.\, Recall that a weight of $1.0$ is just an indication that an assertion exists in ConceptNet \cite{Speer2008AnalogySpaceRT}, so the information we are receiving from ConceptNet in this particular case is pretty weak.\, But nonetheless, we will still make use of that weak information.\, \textsc{Patch-Comm} goes ahead and decides that the ambiguous PP should be attached to the \texttt{play}, and sends that decision back to START.
\par
START resumes its parsing process, only to encounter the second ambiguity.\, So it sends \textsc{Patch-Comm} its partial parse for a second time:
\begin{figure}[!ht]
    \centering
    \begin{verbatim}
            (:ambig-PP
                (PP :prep "with" :det "one" :noun "string")
             :possible-attachments
                ((V :verb "play")
                 (NP :det "the" :noun "guitar")))\end{verbatim}
    \caption{START's 2nd partial parse for sentence (3).}
\end{figure}
\par\noindent
We notice that this is exactly the same scenario as the one described in section 3.1, so everything proceeds in exactly the same way as if they are for sentence (2).\, At the end, START completes its parse result:
\begin{figure}[!ht]
    \centering
    \begin{verbatim}
                [John          play          guitar]
                [play          with          finger]
                [guitar        with          string]
                [string    has_quantity        1   ]
                [finger    has_quantity        1   ]\end{verbatim}
    \caption{START's output for sentence (3), using \textsc{Patch-Comm} querying ConceptNet.}
\end{figure}
\subsection{Using RetroGAN-DRD, A System That Goes Beyond ConceptNet}
The final part of our implementation, as promised, addresses ConceptNet's disconnected-knowledge and insufficient-knowledge problems (section 2.3).\, The solution we have come up with is incorporating the Deep Relationship Discovery (RetroGAN-DRD) system
, a developing system that facilitates commonsense learning and reasoning \cite{ColonHernandez2019DoesAD}.
\par
There are in fact two subsystems: A RetroGAN system, followed by a DRD system.\, The RetroGAN system utilizes the CycleGAN \cite{Zhu2017UnpairedIT} idea to train a target-domain generator that mimicks word vectors that already carry ConceptNet semantics implicitly \cite{speer2016ensemble}.\, Specifically, RetroGAN takes pre-trained fastText \cite{Bojanowski2017EnrichingWV} embeddings in its source domain and retrofitted ConceptNet emsemble \cite{speer2016ensemble} in its target domain, and post-specializes \cite{vulic-etal-2018-post} out-of-knowledge, unseen concepts by learning a generalizable mapping from seen fastText concepts and the retrofitted ConceptNet ensemble.\, The DRD system consists of trained relation-predicting models; each such model corresponds to a ConceptNet relation.\, Essentially, DRD serves to take a pair of post-specialized concepts from RetroGAN, test it on all the relation models, and output the predicted weights for all the models (i.e., relations).
\par
We query the RetroGAN-DRD system in basically the same way that we query ConceptNet itself: Provide two concepts, and ask for the relation that carries the highest weight.
\par
Taking into account both statistical sensitivity and the fact that RetroGAN-DRD is under development, we took some care to ensure better experimental outcomes.\, The first thing we did was, for each of the two concepts, we also introduce $5$ of its neighbors to make a neighborhood of $6$ concepts, represented as a $6 \times 300$ tensor.\, This was done to ensure that ``true'' concepts are not left out due to possible statistical instability of vectorized knowledge.
After drawing two neighborhoods for both concepts, we match up the entire neighborhood of the first concept with that of the second concept, making a total of $6 \times 6 = 36$ pairs.\, Each such pair then gets fed through the RetroGAN-DRD system, and the system outputs the highest-weighted relation along with the weight.\, Overall, we get $36$ relations and $36$ weights.
\par
The second thing we did was that, once all these $36$ weights were collected into a list, we treat the highest $3$ weights and the lowest $3$ weights, along with their corresponding relations, as outliers.\, We remove these outliers to obtain a remainder of $30$ relations and $30$ weights, and we compute the following:
\begin{align*}
    w^{*} = \log\left\{\max_{i=1}^{30}\left\{\text{SOFTMAX}(\exp{(\vec{w})})\right\}\right\}
\end{align*}
where $\vec{w} = \begin{bmatrix} w_{1} & \dotsi & w_{30} \end{bmatrix}$ is the vector of those $30$ weights, and
\begin{align*}
    \exp{(\vec{w})} = \begin{bmatrix}\exp{(w_{1})} & \dotsi & \exp{(w_{30})} \end{bmatrix}.
\end{align*}
This way, $w^{*}$ is a statistically less sensitive measure for how the original pair of concepts drawn from the partial parse should relate.
\par
By way of demonstration, when using RetroGAN-DRD on sentences (1) and (2), we have
\begin{table}[!ht]
    \begin{subtable}[t]{0.45\textwidth}
    \centering
    (1) John is playing the guitar with one finger
    \begin{tabular}{|c|c|c|}
        \hline
        Concept Pair        & Relation      & $w^{*}$ \\
        \hline
        (finger, play)      & SymbolOf      & -3.26578 \\
        (finger, guitar)    & ObstructedBy  & -3.33881 \\
        \hline
    \end{tabular}
    \caption{RetroGAN-DRD's reasoning about the ambiguous PP from sentence (1), refined.}
    \end{subtable}
    \hspace{1.0cm}
    \begin{subtable}[t]{0.45\textwidth}
    \centering
    (2) John is playing the guitar with one string
    \begin{tabular}{|c|c|c|}
        \hline
        Concept Pair        & Relation      & $w^{*}$ \\
        \hline
        (string, play)      & ObstructedBy  & -3.38202 \\
        (string, guitar)    & UsedFor       & -3.37002 \\
        \hline
    \end{tabular}
    \caption{RetroGAN-DRD's reasoning about the ambiguous PP from sentence (2), refined.}
    \label{tab:my_label}
    \end{subtable}
    \caption{RetroGAN-DRD's reasoning about ambiguous PPs.}
\end{table}
\par\noindent
When \textsc{Patch-Comm} uses the information in Tables 4(a) and 4(b), it will still decide that sentence (1) should have ``plays with one finger'' and that sentence (2) should have ``guitar with one string.''\, The only difference is that, by using RetroGAN-DRD, we are now able to overcome both the disconnected-knowledge problem and the insufficient-knowledge problem, at least to some extent.
\par
And finally, START's parse results in this case are:
\begin{figure}[!ht]
    \centering
    \begin{verbatim}
                [John          play          guitar]
                [play          with          finger]
                [string    has_quantity        1   ]\end{verbatim}
    \caption{START's output for sentence (1), using \textsc{Patch-Comm} querying RetroGAN-DRD.}
    \hspace{1.0cm}
    \begin{verbatim}
                [John          play          guitar]
                [guitar        with          string]
                [string    has_quantity        1   ]\end{verbatim}
    \caption{START's output for sentence (2), using \textsc{Patch-Comm} querying RetroGAN-DRD.}
\end{figure}

\newpage
\section{Some Results}
We evaluated our system on section 23 of Belinkov et al.'s English dataset\footnote{We used the publicly available data set from \href{https://github.com/boknilev/pp-attachment/tree/master/data/pp-data-english}{https://github.com/boknilev/pp-attachment/tree/master/data/pp-data-english}.} \cite{belinkov-etal-2014-exploring}.\, Each data point consists of a \textit{preposition}, its \textit{child}, and a set of candidate \textit{heads} to which the PP, consisted of the preposition and the child, is to be attached.
\par
We only evaluated on the preposition ``with'' to be consist with our demonstration examples, and we selected only the data points that have at most 3 candidate heads.\, Overall, there are 17 of them with 2 candidate heads and 31 of them with 3 candidate heads, for a total of 48.\, The statistics of our results are summarized in the following table:
\begin{table}[!ht]
    \centering
    \captionsetup{width=0.9\linewidth}
    \begin{tabular}{|c|c|c||c|}
        \hline
        \textbf{Baseline} & \textbf{Accuracy} & \textbf{$w^{*}$ within $0.1$} & \textbf{Best} \\
        \hline
        39.2\%    &   45.8\%    &   27.1\%    &   \textbf{88.7}\% \\
        \hline
    \end{tabular}
    \caption{Our results on a part of the data set developed by Belinkov et al. 
    Baseline accuracy is computed by assigning $50\%$ to the 17 data points with 2 candidates and $33.333\%$ to the 31 data points with 3 candidates. Accuracy is computed as we got 22 correct out of 48. $w^{*}$ within $0.1$ means, among the data points we got wrong, the $w^{*}$ score for the answer that Patch-Comm picked differs is no more than $0.1$ from the ground-truth answer. The last column is the best result from the paper, obtained from evaluations on their entire data set.}
    \label{table:results}
\end{table}
\par\noindent
These results suggest that the scalability of \textsc{Patch-Comm} awaits.

\section{Discussion and Future Works}
This work is an immediate result of our decision to take on the challenge of scaling up the language-processing capacity of story-understanding systems.\, Scaling-up is important, because all story-understanding systems to date --- with Genesis \cite{genesis-manifesto-2018} being perhaps the most prominent one --- can only read and analyze toy stories, but we want them to start reading and analyzing natural-language stories.\, Scaling-up is a challenge, because natural-language understanding (NLU) is hard.
\par
It soon becomes clear to us that, despite of the seeming bifurcation of \textit{language} and \textit{story} in our setup, both language understanding and story understanding require commonsense knowledge, but perhaps language knowledge and story knowledge should reside on different ``levels.''\, Story knowledge is really \textit{global} information about the story at hand, and what we mean by language knowledge is essentially \textit{local}, sentence-level information.\, Therefore, we think it makes sense in principle for story knowledge to have the ability to override sentence-level language knowledge.\, For example, if there is a story context that tells us that John is a three-year-old kid who plays with all kinds of strings all the time, then all of a sudden, it seems all the more likely that John is actually using a piece of string to play the guitar.
\par
We realize that it is actually important to incorporate story-level knowledge, which a story-understanding system already has, to scale up its sentence-level language understanding.\, Consequently, we have, on our research agenda, the goal of employing global, story-level knowledge to resolve local, sentence-level ambiguities, and to compile the refined local interpretations of all sentences into a global interpretation of the story.\, In fact, a story-understanding system like Genesis is already able to offer global interpretations for toy stories.
\par
Next up, we want to extend our commonsense knowledge-based approach to other challenging linguistic problems, such as co-reference resolution (a.k.a., pronoun disambiguation).\, Our hope for co-reference resolution is to go beyond the state-of-the-art benchmarks, such as the series of works on the Winograd Schema Challenge \cite{levesque-wsc-2011,Sakaguchi2019WINOGRANDEAA,kocijan2020review}, and build programs that can resolve more global co-references, like the ones in the following children's story:
\begin{center}
    \fbox{\begin{minipage}{0.8\textwidth}
        Robbie and Susie were going to Marvin's birthday party. One of them wanted to buy a kite. ``But he already has one,'' he said. ``He will make you take it back.''
    \end{minipage}}
\end{center}
Last but not least, the Holy Grail of NLU is building computer systems that understand discourses.\, To that effect,
story-understanding systems such as the Genesis system are built with the intention of enabling us to gain insights into our own human intelligence, through the lens of story understanding \cite{genesis-manifesto-2018}.\, We hope to scale up to a point where we will gain a sufficient understanding of our own intelligence by looking at the underlying story-understanding processes that are and will continue to be transparent.\, And when we understand and can explain ourselves that much better, we will hopefully have raised our own conception of \textit{trust} to a higher level within society.

\section*{Acknowledgments}
We would like to thank Sue Felshin for sharing with us her valuable experience with and resources of the START natural-understanding system, as well as our in-depth discussions on this project. We would also like to thank Pedro Colón-Hernández for his contibuting the RetroGAN-DRD system as well as numerous insights into the problem of computer language understanding.

\bibliographystyle{coling}
\bibliography{coling2020}

\end{document}